\documentclass[runningheads]{llncs}
\usepackage{graphicx}
\usepackage{amsmath,amssymb} 
\usepackage{color}\usepackage[width=122mm,left=12mm,paperwidth=146mm,height=193mm,top=12mm,paperheight=217mm]{geometry}

\usepackage{epsfig} 
\usepackage{multirow}
\usepackage{epstopdf}
\usepackage{booktabs}
\usepackage{graphicx}
\usepackage{threeparttable}
\usepackage{tabularx}

\usepackage{cite}

\begin{document}
\pagestyle{headings}
\mainmatter

\title{
Leveraging the Power of Gabor Phase for Face Identification: A Block Matching Approach
} 

\titlerunning{Leveraging the Power of Gabor Phase for Face Identification }
\authorrunning{Yang Zhong, Haibo Li}

\author{Yang Zhong, Haibo Li}
\institute{KTH, Royal Institute of Technology}

\maketitle

\begin{abstract}
Different from face verification, face identification is much more demanding.
To reach comparable performance, an identifier needs to be roughly N times better than a verifier. 
To expect a breakthrough in face identification, we need a fresh look at the fundamental building blocks of face recognition. 
In this paper we focus on the selection of a suitable signal representation and better matching strategy for face identification. 
We demonstrate how Gabor phase could be leveraged to improve the performance of face identification by using the Block Matching method. 
Compared to the existing approaches, the proposed method features much lower algorithmic complexity: face images are only filtered by a single-scale Gabor filter pair and the matching is performed between any pairs of face images at hand without involving any training process. 
Benchmark evaluations show that the proposed approach is totally comparable to and even better than state-of-the-art algorithms, which are typically based on more features extracted from a large set of Gabor faces and/or rely on heavy training processes.

\end{abstract}

\section{Introduction}

Face recognition as an active research topic in pattern recognition has been intensively studied for more than two decades \cite{turk1991face, Belhumeur1997, wiskott1997face, ahonen2004face, zou2007comparative, wright2009robust, chan2013multiscale, taigman2013deepface}. 
Face recognition works in two essentially different modes: face verification and face identification. Face verification performs \emph{1:1} matching to provide a binary decision on the claimed identity. The performance of face verification in controlled scenarios has reached a rather high accuracy \cite{givens2013biometric}. 
Recently, more research efforts have been devoted to  face verification in unconstrained conditions. 
By utilizing strong alignment approaches \cite{cao2014face, yi2013towards, chen2012bayesian}, pose-robust matching schemes \cite{hua2009robust, pinto2009far, li2013probabilistic}, and advanced deep-learning techniques\cite{HuangGB2012DL, taigman2013deepface, sun2013deep}, significant improvement in the performance of face verification has been achieved. 
The performance on the benchmark, the Labeled Face in the Wild (LFW) dataset \cite{Huang2007}, is close or even go beyond that of human beings \cite{taigman2013deepface}. 

In contrast, face identification is more difficult.
It performs \emph{1:N} matching to sort out the gallery images based on the pair wise similarity measurement. 
Obviously, the operating requirement in face identification is \textbf{\textit{vastly more demanding}} than operating merely in verification: an identifier needs to be roughly \textbf{\textit{N}} times better than a verifier to achieve comparable odds against making false matches \cite{4052470}. 
This is probably the reason why the progress that has been made in face identification was not significant in the last five years.  
Even though the proposed approaches were more and more complex, the recognition performance on the benchmark evaluation is almost the same \cite{xie2010fusing, yang2013robust, cament2014fusion, 6657757}. 
To make a breakthrough in face identification, it seems that we have to revisit the basis of face recognition, and have a fresh look at the fundamental building blocks of face recognition. 

In face identification, one of the most basic building blocks is the construction of features for measuring the similarity between two face images. 
The construction of features consists of two steps: (1) selection of a suitable representation for face images; (2) feature extraction from the representation. 
There is a large collection of research papers on how to extract stable, local or global discriminative features, 
e.g.\ the commonly used SIFT \cite{lowe2004distinctive}, HOG \cite{dalal2005histograms}, and LBP \cite{ahonen2004face}. 
Recently, these features are criticized because they are `hand-crafted'. 
It was claimed that better features can learned from big data collections through deep-learning approaches \cite{HuangGB2012DL, taigman2013deepface, sun2013deep}. 
But due to inherent complexity of face identification, there is no literature on using deep-learning to learn features for face identification.  
On the other hand, a significant progress has been made in the selection of more suitable representation of faces for face identification. 
Compared to using the raw image pixels, feature extraction from the amplitude or phase spectrum of the Gabor filtered images has shown a significant improvement in the performance of face identification. 
Although the complexity is raised 40 times or even more, the identification rate in benchmark evaluations is improved for more than 20\%, reaching around 90\% \cite{givens2013biometric}.
This is due to the so-called `blessing of dimensionality'.  
Now the question is how to achieve face identification rate from 90\% to 95\%, or even higher.
In this paper we argue that Gabor phase could enable such performance improvement for face identification. 

Most existing face identification algorithms extract discriminative features from the Gabor amplitude. 
The major advantage of the Gabor amplitude is that it varies slowly with the spatial position so that it is robust to texture variations caused by dynamic expressions and imprecise alignment. 
But the problem is Gabor amplitude depends on the imaging contrast and illumination \cite{daugman2001statistical}.
This means that if two photos were taken at two times in different environments (unfortunately, this is the most common practice of face identification that the probe and gallery images were taken in two occasions), it is hard to extract consistent features from the Gabor amplitude. 

Alternatively, the Gabor phase can be utilized since it is robust to light change. 
In fact, it has been well-known for a long time that phase is more important than amplitude from the signal reconstruction point of view (see \cite{oppenheim1981importance}). 
Gabor phase should have played more important role in face identification. 
However, the use of Gabor phase in face recognition is far from being common and successful: it often had worse or nearly the same performance as the amplitude in comparative experiments \cite{gao2006weighted, zhang2009gabor, xie2010fusing, cament2014fusion}. 
This is largely due to two challenging issues: (1) Gabor phase is a periodic function and a hard quantization occurs for every period; 
(2) it is very sensitive to spatial shift \cite{wiskott1997face, zhang2009gabor}, which imposes a rigid requirement on face image alignment. 
The first issue was partly solved by introducing the phase-quadrant demodulation technique \cite{daugman2004iris}, but the second one is still far from being solved. 
The state-of-the-art Gabor phase approach (LGXP \cite{xie2010fusing}) extracts varied LBP from the phase spectrum. 
Since the combination of the phase and LBP is also sensitive to the spatial shift, the power of Gabor phase does not demonstrate in face identification. 

In this paper, we propose a method that merely leverages the power of Gabor phase to conduct face matching for face identification. 
By using the Block Matching method \cite{7025145}, our approach does not depend on training process or fusion of any other features.
Benchmark evaluations on the FERET show that the identification rate reaches 95\% on the hardest `Dup2', which is the best result reported in the literature so far.


\section{Related work}

In this section, we first briefly review the Gabor representation and recent Gabor based methods that utilizing the Gabor amplitude or phase in different ways. 
Then we introduce the Block Matching method, which features a different matching strategy from the most existing approaches, used in this work.

\subsection{Gabor Wavelet and Related Approaches}
A Gabor face is obtained by filtering the image with the Gabor filter function, which is defined as:
\begin{equation}
\nonumber
\psi _{u,v}(z) = \frac{\left \| k_{u,v} \right \|^2}{\sigma ^2} 
e^{(-\left \| k_{u,v} \right \|^2 \left \| z \right \|^2 / 2\sigma ^2)} 
[ e^{i k_{u,v} z} - {e^{-\sigma ^2 /2}} ] ,
\end{equation} 
where $u$ and $v$ define the orientation and scale of the Gabor kernels respectively,
and the wave vector is defined as: 
\begin{equation}
\nonumber
k_{u,v} =  k_{v} e^{i\phi _{u} },
\end{equation} 
where $ k_{v} = k_{max} / f^{v}$, $\phi _{u} =u \pi /8 $; 
$k_{max}$ is the maximum frequency, $\sigma$ is the relative width of the Gaussian envelop, 
and $f$ is the spacing factor between kernels in the frequency domain \cite{liu2002gabor}. 
The discrete filter bank of 5 different spatial frequencies ($v \in [0,\cdots ,4]$) and 8 orientations ($u \in [0, \cdots ,7]$) 
are mostly exploited to filter the face images to facilitate multi-scale analysis. 

To extract the Gabor features, one popular way is to extract the LBP-type patterns from the complex Gabor transformed image. 
As in \cite{zhang2005local}, the LGBP feature is extracted from the amplitude spectrum.
In \cite{4032834} and \cite{xie2010fusing}, 4-ray phase-quadrant demodulator is applied to demodulate the phase from each complex Gabor coefficients, and local binary phase descriptors are subsequently generated from the demodulated phase spectrum.  
Gabor local descriptors can also be built by applying dimension reduction technique (e.g.\ FLD) to the raw Gabor amplitude patches as in \cite{su2009hierarchical}, or by computing the local representation exemplified as GOM \cite{6657757} and SLF \cite{yang2013robust}. 

Fusing the features that are independent of the local Gabor features can also lead to better performance: 
\cite{tan2007fusing, su2009hierarchical, 4032834} fuses the global (holistic) features with local ones on feature level;
\cite{xie2010fusing} proposes fusion of Gabor phase and amplitude on the score and feature levels;  
\cite{6657757} fuses real, imaginary, amplitude and phase. 
Alternatively, attaching illumination normalization step and weighting the local Gabor features is shown to be helpful as well \cite{cament2014fusion}.

\subsection{The Block Matching method}
While most local matching approaches perform matching only between the spatially corresponding patches, some others allow each segmented patch of one image to search the best matching from spatially neighboring locations on the other image to achieve better robustness to the spatial shift of the textures as in \cite{lades1993distortion, wiskott1997face, hua2009robust, 7025145}. 

The Block Matching method proposed by [Anonymous authors] \cite{7025145} exploits the searching strategy to find best patch-wise matches, and during the same process, performs pair-specific normalization to compute the patch-wise distance. 
It can conduct face matching with any two face pair at hand without training.
Although only raw pixel intensity was used, the evaluations show that the Block Matching outperformed the LBP method and its improved high-order variants \cite{7025145}.

\section{The Gabor Phase Block Matching Approach}

Although feature descriptors and similarity metrics are different, most of the existing local based methods perform matching of local features only between spatially corresponding patches. 
This means that the matching philosophy embedded in most of the local based approaches by default sees the spatially corresponding patches/features as the best match (since they perform matching only between spatially corresponding local regions).
It might be true when two faces have different shapes and configurations, but  due to the movement of facial components, head pose or imprecise alignment, the spatially corresponding patches would cover different face regions (as Fig. 2 in \cite{zou2007comparative}).
That is, due to the sensitiveness to the spatial shift of Gabor phase, the best matching regions are likely located away from the corresponding locations even if the matching face pairs are aligned. 
A better solution is to utilize searching scheme that enable each region to search for the best matching.
In other words, to fairly match the local phase descriptors it is intuitive to combine a searching strategy that enables each local phase patch to search for its best match from the neighborhood locations.

\begin{figure*}[t]
\begin{center}
   \includegraphics[width=0.9\linewidth]{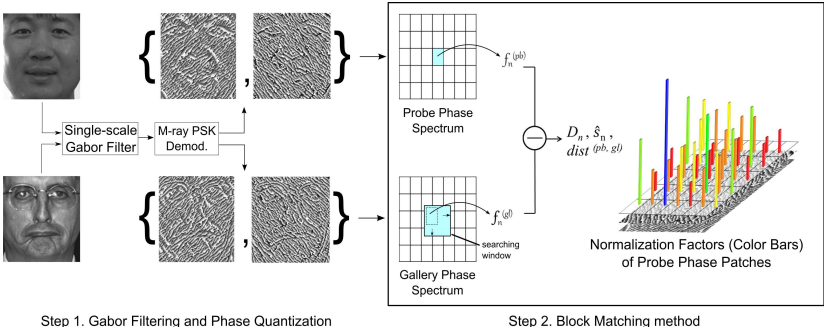}
\end{center}
   \caption{Gabor Phase Matching using the Block Matching method}
\label{fig:phaseMatch}
\end{figure*}

\begin{figure}[t]
\begin{center}
   \includegraphics[width=0.5\linewidth]{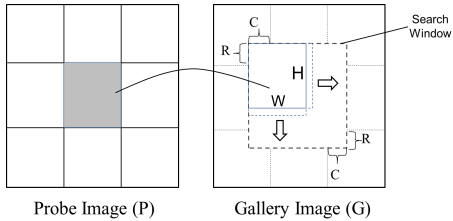}
\end{center}
\vspace{-1.2em}
   \caption{ The Block Matching scheme}
\label{fig:blkmatch}
\vspace{-0.5em}
\end{figure}

Based on this thinking, we adopt the Block Matching method \cite{7025145} for matching Gabor phase . 
The matching process of our Gabor Phase Block Matching (GPBM) is illustrated in Fig.\ref{fig:phaseMatch} with the details of the Block Matching searching scheme shown in Fig.\ref{fig:blkmatch}.
A two-step matching process is conducted for matching a probe-gallery face pair.
First, the face images are transformed to Gabor space using two Gabor filters of 1 scale and 2 orientations.
The filtered images are demodulated by a Gray-coded Phase Shift Keying (PSK) demodulator to quantize the phase. 
Second, the demodulated phase spectra are inputted to the Block Matching method \cite{7025145} to determine the pair wise distance between a probe ($pb$) image and a gallery ($gl$) image.

Specifically in Step 2 of Fig. \ref{fig:phaseMatch}, we first segment the probe phase spectrum into $N$ non-overlapping patches and each of the patches $\{f^{(pb)}_{n}\}_{0}^{N-1}$ is simply formed by the \textit{\textbf{raw phase}} code of the block.
For each probe patch $ f^{(pb)}_{n} $ centered at image coordinate $(x_{n}, y_{n})$ (noted as $ f^{(pb)}(x_{n}, y_{n}) $), 
it searches within the corresponding searching window and yields a patch-wise distance vector $D_{n}$ which is noted as:
\begin{equation}
D_{n} = \{ d_{i}^{n} \}, i\in [0, L-1]
\end{equation}
where $L$ is the number of candidate gallery patches within the $(2R+1)\times (2C+1)$ searching window, i.e.\ $ L = (2R+1)\cdot (2C+1)$ when applying full search method, $R$ and $C$ stands for the searching offset in vertical and horizontal directions respectively. 
Each element in $D_{n}$ is computed by performing explicit matching over the raw demodulated phase as:
\begin{equation}
d_{i}^{n} = \left \| \textrm{XOR}(f^{(pb)}(x_{n}, y_{n}) , f^{(gl)}(x_{i}, y_{i})  )_{decimal} \right \|_{2},
\end{equation}
where the patch-wise distance metric is the 2-norm of element wise Humming distance in decimal 
and $(x_{i}, y_{i})$ denotes the coordinate of the patch center within searching window on the gallery face image so that,
\begin{equation}
\begin{cases}
x_{i} = x_{n} + \Delta x, & \Delta x \in [-C, C]  \\ 
y_{i} = y_{n} + \Delta y, & \Delta y \in [-R, R].
\end{cases}
\end{equation}

We then calculate the slope $k_{n}$ of the linear fitting of the first 5 ascendingly sorted values of $D_{n}$ for normalization of the patch wise distance for each patch, 
such that the preliminary normalization factor $s_{n}$ is
\begin{equation}
s_{n} = k_{n} / d_{n},
\end{equation}
where $d_{n} = min(D_{n})$. 
It is further normalized as:
\begin{equation}
\hat{s}_{n} = s_{n} / \sum_{n=0}^{N-1} s_{n}.
\end{equation}
Finally, the distance between a matching pair of probe and gallery face is
\begin{equation}
dist^{(pb, gl)} = \sum_{n = 0}^{N-1} ( \hat{s}_{n} \cdot d_{n} ).
\end{equation}


\section{Experiments and Results}
\subsection{Database Selection}
There are a variety of large-scale datasets available for benchmark evaluation of different face recognition approaches, such as the FERET \cite{phillips2000feret}, FRGC 2.0\cite{phillips2005overview} and the LFW dataset.
But the FRGC2.0 and LFW are dedicated face verification benchmarks. 
Thus we select the FERET database which is the most commonly recognized face identification benchmark to evaluate and compare our method with state-of-the-art approaches. 
In addition, the CMU-PIE \cite{sim2002cmu} dataset is selected to evaluate our GPBM under variations of pose, expression and illumination in the probe images.
%
 
\subsection{Experimental Setup}
\begin{figure}[t]
\begin{center}
   \includegraphics[width=0.5\linewidth]{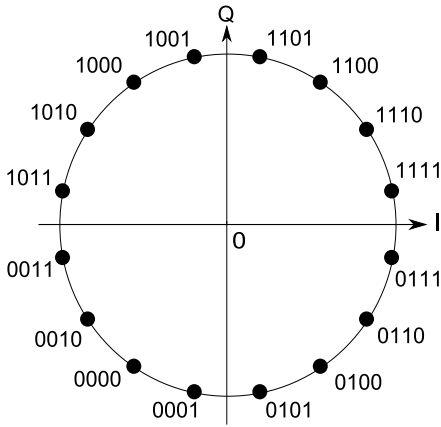}
\end{center}
   \caption{16-PSK Demodulator Constellation}
\label{fig:demod}
\end{figure}

Face images are first normalized based on the positions of both eyes and the Gabor filtered face image to $150 \times 136$ so that the same height/width ratio (1.1 : 1) is maintained the same as in \cite{xie2010fusing, zhang2010local}. 
We use a $6\times6$ Gabor filter with $v = 0$, $u \in \{2,6\}$, $f = \sqrt{2}$, $k_{max} = \pi/2$, $\sigma = 2\pi$. 
Accordingly, Gray-coded 16-PSK demodulator is used for phase quantization and the constellation is shown in Fig.\ref{fig:demod}.  

\subsection{Evaluations on the CMU-PIE database}

The CMU-PIE database contains 41368 images of 68 subjects.
Images with Pose Label 05, 07, 09, 27, and 29 under 21 illuminations (Flash 2 to 22) of all the 68 persons are selected as the probe set.

\begin{figure}[t]
\begin{center}
   \includegraphics[width=0.5\linewidth]{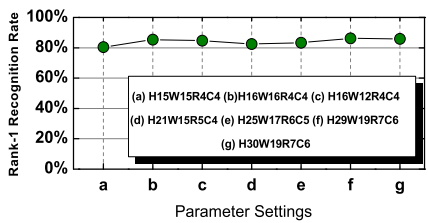}
\end{center}
\vspace{-2em}
   \caption{Recognition Rates Under Different Parameters}
\label{fig:newPIEpara}
\end{figure}

In the Blocking Matching method, the most important parameters are the block size ($H$ and $W$) and searching offset ($R$ and $C$). 
We have conducted a set of empirical tests over other datasets to select suitable parameters. 
We found that $H$ and $W$ satisfying $R \approx max(H, W)/4$, $R/C \approx 1.2$ and $H/W \approx 1.5$ work quite good. 
In our experiments, we select the block size of $29 \times 19$ and search offset of $R = 7$, $C = 6$ pixels (a tradeoff between performance and complexity). 
For this group of parameters we selected the first 2000 probe images to evaluate its effect on the performance. 
In addition, we check how sensitive the performance is to the selection of parameters by looking around the selected group of parameters. 
The test results are shown in Fig. \ref{fig:newPIEpara}. 
From the test results one can see that the performance is quite robust to the selection of parameters.

We then conduct experiments on the CMU-PIE probe set and compare our GPBM with G\_LBP and G\_LDP \cite{zhang2010local}.
The G\_LBP is the Gabor version LBP and the G\_LDP is a type of improved Gabor amplitude Local Binary Pattern.
The G\_LDP achieved equivalent performance as LGXP (Gabor Phase pattern) on the FERET evaluations so that it is a good reference for comparison. 
The comparative rank-1 recognition rates are listed in Table \ref{table:cmupie}.
It can be seen that our method is at least 3\% better than the G\_LDP, even though LDP extract much more effective patterns than the LBP from the Gabor amplitude space.
Utilizing the Gabor phase in the Block Matching scheme is more effective in dealing with pose and illumination changes than LBP-type patterns extracted from the Gabor amplitude space. 

\begin{table}
\centering
\caption{Comparative rank-1 recognition rates of GPBM on the CMU-PIE database}
\label{table:cmupie}
\begin{tabular}{@{}lcc@{}}
\toprule
Method & Accuracy \\ \midrule
G\_LBP $^{\ast}$								& 71\% \\
G\_LDP 2nd-order $^{\ast}$ 		& 72\% \\
G\_LDP 3rd-order $^{\ast}$			& 79\% \\
G\_LDP 4th-order $^{\ast}$			& 74\% \\ 
\cmidrule{1-2}
\textbf{GPBM} 							& \textbf{82\%} \\
\bottomrule
\end{tabular}
\begin{tablenotes}
  \item $^{\ast}$ \footnotesize{The recognition rates are estimated from Fig.\ 12a in \cite{zhang2010local}}. 
\end{tablenotes}
\end{table}

\subsection{Evaluations on the FERET database}

The FERET database contains 1196 frontal images in the gallery set, 1195 images with different expressions in the probe set `Fa', 
194 images with illumination variations in the probe set `Fc', 722 images taken in later time in the `Dup1' set, and 234 images taken at least 1 year later than the
gallery set form the hardest `Dup2' set. 
We faithfully follow the evaluation protocol of the FERET dataset. 
The results of our GPBM with other approaches using Gabor-phase are listed in Table \ref{table:GPVs}.

\begin{table}
\centering
\caption{Comparative rank-1 recognition rates of \textbf{Gabor-phase} based approaches on the FERET database}
\label{table:GPVs}
\begin{tabular}{@{}lcccc@{}}
\toprule
Method                             						& Fb           & Fc          & Dup1       & Dup2          	\\ \midrule
LGBP\_Pha  \cite{zhang2009gabor}      			& 93\%         & 92\%        & 65\%       & 59\%        \\
LGBP\_Pha$^{weighted}$ 	\cite{zhang2009gabor}	& 96\%         & 94\%        & 72\%       & 69\%        \\
HGPP$^{weighted}$	\cite{4032834}							&  97.5\%      &  99.5\%     & 79.5\%     & 77.8\%        \\
LGXP		\cite{xie2010fusing}									& 98\%         & 100\%       & 82\%       & 83\%        \\
LGXP+BFLD	\cite{xie2010fusing}		     				& 99\%         & 100\%       & 92\%       & 91\%        \\ 
S[LGBP\_Mag+LGXP] \cite{xie2010fusing} 			& 99\%         & 100\%       & 94\%       & 93\%        \\ 
LMGEW//LN+LGXP \cite{cament2014fusion}				& 99.9\%         & 100\%       & 94.7\%       & 91.9\%        \\ 

\cmidrule{1-5}
\textbf{GPBM}									& \multicolumn{1}{l}{\textbf{99.4\%}} 	& \multicolumn{1}{l}{\textbf{100\%}} & \multicolumn{1}{l}{\textbf{95.3\%}} 			& \textbf{94.9\%} \\
\bottomrule
\end{tabular}
\end{table}


From Table \ref{table:GPVs} we can see that in a fair comparison, where only Gabor phase is utilized for matching, our GPBM is almost 12\% better than LGXP in the hardest `Dup2'; 
even in unfavorable comparisons, where pre-processing, training, and fusion method are exploited by LMGEW//LN+LGXP and S[LGBP Mag+LGXP], our GPBM still outperforms.
To our best knowledge, the S[LGBP\_\ Mag+LGXP] --- aided by the Gabor amplitude and training procedures --- was state-of-the-art Gabor phase based method in the hardest FERET `Dup2', and our GPBM is better than that.

\begin{table}[h]
\centering
\caption{Comparative Summary with Recent State-of-the-art Face Identification Approaches}
\label{table:summaryAlgos}
\resizebox{\textwidth}{!}{\begin{tabular}{ lcccccc}
\toprule 
\multicolumn{1}{c|}{Methods} & \multicolumn{1}{c|}{Image Size} & \multicolumn{1}{c|}{\begin{tabular}[c]{@{}c@{}}Photometric\\ Processing\end{tabular}} & \multicolumn{1}{c|}{\begin{tabular}[c]{@{}c@{}}Gabor Filter\\ Scale Orient.\end{tabular}} & \multicolumn{1}{c|}{\begin{tabular}[c]{@{}c@{}}Gabor Feature\\ Space\end{tabular}} & \multicolumn{1}{c|}{\begin{tabular}[c]{@{}c@{}}Training Data\\ Independent\end{tabular}} & \begin{tabular}[c]{@{}c@{}}Rank-1 Rate\\ on FERET Dup2\end{tabular} \\ \midrule
    LGXP  \cite{xie2010fusing}   		 						&$88 \times 80$ 	& No 		& $5 \times	8$ & Phase 									& No   & 83\% 	\\
		S[LGBP+LGXP]	 \cite{xie2010fusing}		&$88 \times 80$ 	& No		& $5 \times	8$ & Amplitude + Phase 			& No   & 93\% 	\\ %
		GOM	\cite{6657757}													&$160 \times 128$ & No 		& $5 \times	8$ & Amplitude + Phase 			& No   & 93.1\% \\ 
		LN+LGXP \cite{cament2014fusion}			&$251 \times 203$ & Yes 	& $5 \times	8$ & Phase 									& No   & 91.9\% \\
		LN+LGBP \cite{cament2014fusion}			&$251 \times 203$ & Yes 	& $5 \times	8$ & Amplitude							& No   & 93.6\% \\

		SLF-RKR\_\textit{l}2 \cite{yang2013robust}	&$150 \times 130$ & No 		& $5 \times	8$ & Amplitude 							& No   & 94.4\% \\ 
		\cmidrule{1-7}
\textbf{GPBM} & $ 150 \times 136 $  & No & $1 \times 2$& \begin{tabular}[c]{@{}c@{}}Phase\\ (\textbf{explicit matching})\end{tabular} & \textbf{Yes} & \textbf{94.9\% }\\ 
\bottomrule 
\end{tabular}}
\end{table}

We also comprehensively compare our GPBM with other recent state-of-the-art approaches based on other techniques on the FERET in Table \ref{table:summaryAlgos}.
From the table one can see that all these approaches are based on Gabor features, which indicates the Gabor space is a very effective signal representation.
Our GPBM method outperforms all the other approaches on the hardest `Dup2' set and it features three advantages: 
1) it only requires two Gabor filters, which is 20 times less than other methods which utilized 40 Gabor filters; 
2) it does not require any training procedure or any attached database to perform face matching;
3) it uses merely raw phase for matching and no feature extraction was performed.

The computational complexity is always a big concern. 
By Table 4 in \cite{mu2011shift}, under the image size of $128 \times 128$ with a $5 \times 8$ Gabor filter bank, the histogram extraction of LGBP takes around 0.45 second, S[LGBP\_\ Mag+LGXP] takes 0.99 second.
Extracting GOM feature takes 0.7 second \cite{6657757}. 
But for our method, the `feature extraction' time is 0 since only raw phase is used for matching; the demodulation is the only required on-line computation of the probe face and it is extremely fast.
Our Matlab implementation executes the matching of a face pair in 0.05 second in average (Gabor filtering included) on a 3.4GHz Intel CPU.
We can therefore safely conclude that our GPBM outperforms the best Gabor-phase based approach (S[LGBP\_\ Mag+LGXP]) in efficiency for a big margin and we can also infer that the other methods in Table \ref{table:summaryAlgos} could hardly be more efficient than our GPBM due to higher image resolution, Gabor face dimensions, and additional photometric processing.
Here we should mention that our GPBM used the very basic `exhaust search' strategy which is not an efficient choice for the matching.
Our preliminary experiments show that there is a large potential to achieve even higher efficiency by using optimized searching strategy.
(Optimization of the Block Matching framework is outside the scope of this paper.)    

\section{Conclusions}

In this paper, we propose a plain approach to leverage the demodulated Gabor phase for face identification based on the Block Matching method. 
The proposed approach dose not utilize a large Gabor filter bank or any training process. 
Rather, it only depends on the signal representation from a single-scale Gabor filter pair and performs explicit matching over the raw Gabor phase spectrum.

Comparative experiments show that: 1) our approach features the highest accuracy utilizing the Gabor phase for face recognition; 2) our approach has very low computational complexity with totally comparable performance to state-of-the-art methods. 
Our experiments also confirm that the Gabor phase is a powerful source to construct features for face identification. 
To leverage the power of the Gabor phase, the key is to have a suitable feature construction approach. 
In this paper, we show that the Blocking Matching is a good choice for such purpose. 
We strongly recommend the Block Matching as an alternative to the commonly adopted LBP should be used more in face recognition.

\clearpage

\bibliographystyle{splncs03}
\bibliography{egbib}
\end{document}